\documentclass{article}
\usepackage{spconf,amsmath,graphicx}
\usepackage{cite}
\usepackage{amsmath,amssymb,amsfonts}
\usepackage{algorithmic}
\usepackage{graphicx}
\usepackage{tabularx}
\usepackage{multicol}
\usepackage{diagbox}
\usepackage{textcomp}
\usepackage{booktabs}
\usepackage{multirow}
\usepackage{hyperref}
\hypersetup{
    colorlinks=true,
    linkcolor=blue,
    filecolor=blue,      
    urlcolor=blue,
    citecolor=cyan,
}
\usepackage{subfigure}
\usepackage{xcolor}
\usepackage{bm}
\usepackage{color}

\title{Exploring the Benefits of Differentially Private Pre-training and Parameter-Efficient Fine-tuning for Table Transformers
}
\name{Xilong Wang$^\star$, Chia-Mu Yu$^\dagger$, and Pin-Yu Chen$^\ddagger$}
\address{$^{\star}$ 
University of Science and Technology of China, Hefei, China\\
$^\dagger$ 
National Yang Ming Chiao Tung University, Hsinchu, Taiwan\\
$^\ddagger$ IBM Research, New York, USA}
\begin{document}
%
\maketitle
\begin{abstract}
For machine learning with tabular data, Table Transformer (TabTransformer) is a state-of-the-art neural network model, while Differential Privacy (DP) is an essential component to ensure data privacy. In this paper, we explore the benefits of combining these two aspects together in the scenario of transfer learning -- differentially private pre-training and fine-tuning of TabTransformers with a variety of parameter-efficient fine-tuning (PEFT) methods, including Adapter, LoRA, and Prompt Tuning.
Our extensive experiments on the ACSIncome dataset
 show that these PEFT methods outperform traditional approaches in terms of the accuracy of the downstream task and the number of trainable parameters, thus achieving an improved trade-off among parameter efficiency, privacy, and accuracy. Our code is available at \url{https://github.com/IBM/DP-TabTransformer}.
\end{abstract}

\section{Introduction}
Table transformer (TabTransformer) \cite{huang2020tabtransformer} is a novel deep tabular data modeling for various scenarios, such as supervised and semi-supervised learning. Its main contribution is to transform regular categorical embeddings into contextual ones, thus achieving higher accuracy compared to previous state-of-the-art methods. On the other hand, differential Privacy (DP) \cite{dwork2006calibrating} is a frequently used technique to ensure privacy for individual data points in a training dataset. DP-SGD \cite{abadi2016deep}, which combines DP with stochastic gradient descent (SGD), is one of the most frequently used optimization techniques in machine learning (ML) to train models on sensitive data while safeguarding individual privacy.

In the literature, DP-SGD techniques either fine-tune a pre-trained model or train a model from scratch. However, almost none of them have focused on TabTransformer.
In this paper, we implement various recent parameter-efficient fine-tuning techniques, such as LoRA \cite{hu2021lora}, Adapter \cite{houlsby2019parameter}, and Prompt Tuning \cite{jia2022visual} (both Shallow Tuning and Deep Tuning), so as to explore the benefits of differentially private pre-training and fine-tuning for TabTransformers. To summarize, our key contributions are as follows: 1) We study an unexplored scenario for transfer learning in TabTransformers with DP, i.e., implementing various kinds of parameter-efficient techniques in the fine-tuning stage instead of full tuning. 
2) Different from previous tabular learning methods which mainly exploited DP at the fine-tuning stage, we study the use of DP-SGD for both pre-training and fine-tuning, thus ensuring end-to-end privacy.
3) Our experiments on the ACSIncome dataset showed that the accuracy outperforms the baselines in most cases, while the parameter efficiency improves by more than $\textbf{97.86\%}$. In addition, we report the best advantageous PEFT setting to inform and inspire the future design of DP-aware pre-training and fine-tuning for TabTransformers.


\section{Background and Related Works}
\subsection{Differential Privacy (DP) and DP-SGD}
ML is widely known for its ability to analyze large datasets, identify patterns, and make predictions or decisions based on that data. However, this also introduces the risk of disclosing sensitive information from the training dataset. DP \cite{dwork2006calibrating} and DP-SGD \cite{abadi2016deep} are introduced to address this issue. A randomized algorithm $\mathcal{A}$ satisfies $(\epsilon,\delta)-DP$ if it holds that:
\begin{equation}
    \mathcal{P}[\mathcal{A}(\mathcal{D})\in S] \leq \exp(\epsilon) \mathcal{P}[\mathcal{A}(\mathcal{D}')\in S] + \delta,
\end{equation}
where $\mathcal{P}[\mathcal{A}(\mathcal{D})\in S]$ is the probability that the output of $\mathcal{A}$ on dataset $\mathcal{D}$ falls within a set $S$, and $\mathcal{P}[\mathcal{A}(\mathcal{D}')\in S]$ is the probability that the output of the $\mathcal{A}$ on a neighboring dataset $\mathcal{D}'$ (which differs from $\mathcal{D}$ by one data point) falls within $S$. The smaller $\epsilon$ is, the stronger privacy guarantee $\mathcal{A}$ has.

Inspired by DP, Differential Privacy Stochastic Gradient Descent (DP-SGD) \cite{abadi2016deep} is one of the most widely used privacy-preserving optimization techniques in ML \cite{kim2021federated,dupuy2022efficient,arif2023reprogrammable,li2023exploring}. It is a two-stage procedure. Formally, given the SGD gradient estimator $g$ evaluated
on the training dataset, and define its sensitivity $S_g$ as the maximum
of $ \left\|g(\mathcal{D}) - g(\mathcal{D}')\right\|_2 $. In the first stage, DP-SGD adds a zero-mean Gaussian noise with a given covariance matrix, i.e., $\mathcal{N}(0, S^2_g\sigma^2\text{\textbf{I}})$ to the computed gradient estimator as follows:
$$g(\mathcal{D})+\mathcal{N}(0,S^2_g\sigma^2\text{\textbf{I}}).$$
In the second stage, DP-SGD passes the gradient estimator through the Clip operator:
$$\text{Clip}(x) = \frac{x}{\max\{1,\left\|x\right\|_2/C\}},$$
so as to fix the sensitivity of the gradient estimator at a hyperparameter $C$. However, regrettably, almost none of DP-SGD techniques have been applied to the study of TabTransformer.

\subsection{Parameter-Efficient Fine-Tuning (PEFT)}
PEFT \cite{pfeiffer2020mad,mao2022unipelt,li2021prefix,houlsby2019parameter,hu2021lora,jia2022visual} is an emerging technique in the field of transfer learning that aims to adapt large pre-trained models to specific tasks with a smaller number of task-specific parameters. It fine-tunes the pre-trained model on a target task while keeping the majority of the original model's parameters frozen. Compared to full-tuning which fine-tunes the entire model, this approach reduces the computational resources and memory requirements needed for task-specific adaptation. PEFT is particularly valuable in scenarios with limited computing resources or when deploying models to resource-constrained environments, without sacrificing task performance. The most popular PEFT techniques include LoRA \cite{hu2021lora}, Adapter \cite{houlsby2019parameter}, and (Deep/Shallow) Prompt Tuning \cite{jia2022visual}. Nevertheless, similar to standard ML, PEFT also faces the risk of disclosing sensitive data throughout the fine-tuning procedure and thus needs a privacy guarantee \cite{yu2021differentially}.

\section{Methodology}

\subsection{TabTransformer}
\begin{figure}[!t]
\centering
\subfigure[Architecture of TabTransformer.]{
		\includegraphics[scale=0.7]{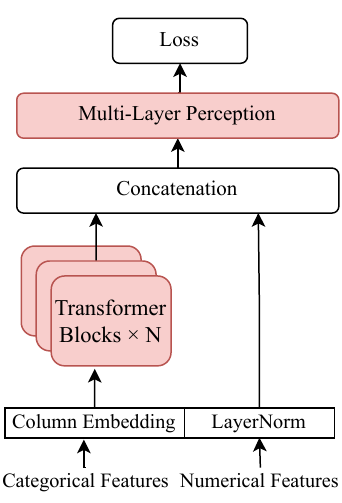}}\hspace{1mm}
\subfigure[Overview of Deep Tuning and Shallow Tuning.]{
		\includegraphics[scale=1.05]{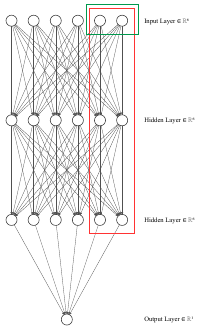}}
\caption{Genreral framework of parameter-efficient tuning on TabTransformer.}
\label{fig1}
\end{figure}

TabTransformer \cite{huang2020tabtransformer} is a deep learning architecture for tabular data modeling. It uses contextual embeddings to achieve higher prediction accuracy and better interpretability. It outperforms state-of-the-art deep learning methods for tabular data and is highly robust against missing or noisy data features. The brief structure of TabTransformer is displayed in Fig. \ref{fig1} (a). The TabTransformer architecture consists of a column embedding layer, a stack of $N$ Transformer blocks, and a multi-layer perceptron (MLP). Each Transformer layer comprises a multi-head self-attention layer followed by a position-wise feed-forward layer. As shown in Fig. \ref{fig1} (a), the areas highlighted in red are where we can perform PEFT. To be specific, we implemented LoRA \cite{hu2021lora} and Adapter \cite{houlsby2019parameter} in Transformer Blocks, while Deep Tuning and Shallow Tuning \cite{jia2022visual} were exploited in MLP.
\subsection{Deep Tuning and Shallow Tuning}
Visual Prompt Tuning (VPT) \cite{jia2022visual} is an efficient alternative compared to full fine-tuning for large-scale Transformer models. It offers two tuning strategies: VPT-Deep and VPT-Shallow. VPT-Deep prepends a set of learnable parameters to each Transformer encoder layer’s input, while VPT-Shallow only inserts the prompt parameters to the first layer’s input. Inspired by VPT, we proposed Deep Tuning and Shallow Tuning, which aims at fine-tuning the MLP of TabTransformer.

\subsection{Adapter}
Adapter \cite{houlsby2019parameter}, as shown in Fig. \ref{fig2} (b), is a transfer learning approach that allows for efficient parameter sharing and extensibility in large pre-trained models. It uses small and task-specific modules that are inserted between the pre-trained layers of the base model. These modules have a near-identity initialization and a small number of parameters, which allows for stable training and slow growth of the total model size when more tasks are added.

\subsection{LoRA}
LoRA \cite{hu2021lora}, as shown in Fig. \ref{fig2} (c), is a low-rank adaptation technique that reduces the number of trainable parameters for downstream tasks while maintaining high model quality. It works by injecting a low-rank adaptation matrix into the pre-trained model, which can be shared and used to build many small LoRA modules for different tasks. LoRA makes training more efficient and allows for quick task-switching.

\subsection{Joint PEFT with DP-SGD}
\begin{figure}
    \centering
    \includegraphics[width = \linewidth]{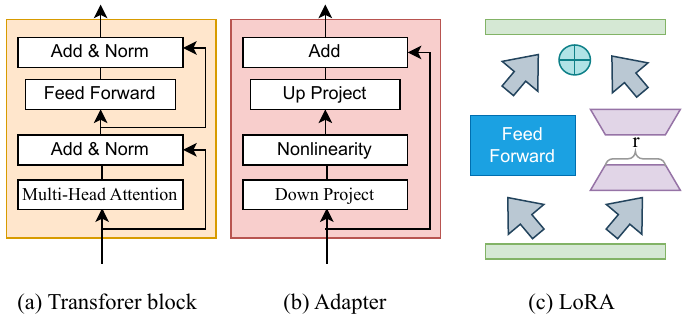}
    \caption{PEFT techniques applied on Transformer block.}
    \label{fig2}
\end{figure}

We incorporate DP-SGD into PEFT by initially pre-training a TabTransformer model with DP on a pre-training dataset. Subsequently, we freeze the backbone of the pre-trained model and apply the aforementioned PEFT techniques to fine-tune the model in conjunction with DP-SGD on the downstream dataset. This approach serves to safeguard the privacy of both the pre-training dataset and the downstream dataset, thus ensuring end-to-end privacy. To be more detailed, as shown in Fig. \ref{fig2} (a) and Fig. \ref{fig2} (c), we combine LoRA with Feed Forward Layer in each Transformer block of TabTransformers. Moreover, we inject an Adapter between the Feed Forward Layer and the Add \& Norm Layer in each Transformer block of TabTransformers. For Deep Tuning and Shallow Tuning, as shown in Fig. \ref{fig1} (b), Deep Tuning tunes certain neurons in every layer of MLP, i.e., the red-marked part in Fig. \ref{fig1} (b). Meanwhile, Shallow Tuning only tunes a few neurons in MLP's input layer, i.e., the green-marked part in the figure.


\section{Performance Evaluation}
In this section, we test the performance of all mentioned PEFT approaches and identify the most effective one to benefit future research. For comparison, we exploit two baselines, i.e., full tuning and training from scratch. Furthermore, to illustrate the impact of PEFT, we also evaluate the pre-trained model directly on the downstream data without PEFT (i.e., Zero-shot Inference). Experimental results clearly show that PEFT methods ensure high parameter efficiency without the loss of accuracy, thus outperforming basic approaches in terms of accuracy, parameter efficiency, and privacy.
\subsection{Experiment Setup}

\textbf{ACSIncome dataset}. The ACSIncome dataset \cite{ding2021retiring} is derived from the American Community Survey (ACS) Public Use Microdata Sample (PUMS) data. It aims to predict whether US working adults' yearly income is above \$50,000. It covers all states of the United States and includes multiple years. Inspired by this feature, we can perform pre-training and fine-tuning across different states. To be more detailed, we chose two states, California (CA) and Indiana (IN), which are geographically distant and have significant differences in population size and economic disparities. CA and IN exhibit obvious training set distribution shifts, and therefore we utilized them for the study of pre-training and fine-tuning with TabTransformer. 

\textbf{Baselines}. (1) Full Tuning: In this scenario, after pre-training, Full Tuning tunes all parameters of the pre-trained model. (2) Train from scratch: This baseline simply trains the entire model on the downstream dataset from scratch without pre-training. (3) Zero-shot Inference: To emphasize the effect of PEFT, after obtaining a pre-trained model, this baseline directly evaluates the performance of the same model on the downstream dataset.

\textbf{Parameters}. (1) Let $e_p$ indicate $\epsilon$ used for pre-training and $e_f$ denote $\epsilon$ used for fine-tuning. The values for $e_p$ and $e_f$ are chosen from $\{0.5,1,2,4,8,16,32\}$. (2) Clipping norm $C=2$. (3) $\delta=10^{-5}$. (4) For TabTransformer, we set the hidden (embedding)
dimension, the number of Transformer blocks, and the number of attention heads to be 32, 4, and 8, respectively. The size of MLP is 5 layers with 72 units for each layer. (4) Batch Size $\mathcal{B}=64$ for both pre-training and fine-tuning. (5) Full tuning tunes 8 units (tokens) in every MLP layer, and Shallow Tuning tunes 8 units just in the first layer of MLP.

\subsection{Experimental Results}
\textbf{Number of Trainable Parameters.} The degree of parameter efficiency in a PEFT technique hinges on the number of parameters that remain trainable during fine-tuning. Let $N$ represent the number of trainable parameters, and then the $N$ of all the techniques are shown in Table \ref{tab1}.
\begin{table}[htbp]
\vspace{-4mm}
\centering
\caption{Number of Trainable Parameters of Various Methods}
\resizebox{\linewidth}{!}{
    \begin{tabular}{|c|c|c|c|}
    \hline
        Methods & Deep Tuning & Full Tuning & Shallow Tuning \\
    \hline
      $N$ & 4,408  & 206,193 &  2,072    \\
    \hline
        Methods & Adapter & LoRA & Train from Scratch \\
    \hline
      $N$  & \textbf{1,424} & \textbf{1,424} & 206,193    \\
    \hline
    \end{tabular}
    }
\label{tab1}
\end{table}

\begin{table}[htbp]
\centering
\caption{Testing Accuracy Comparison.}
\resizebox{\linewidth}{!}{
\begin{tabular}{@{}|c|ccccccc|@{}}

\hline
\multicolumn{8}{|c|}{{Methods}} \\
\bottomrule
\toprule
\multicolumn{8}{|c|}{Full Tuning} \\
\hline
                               \multicolumn{1}{|c|}{\diagbox{$\epsilon_{p}$}{$Acc$}{$\epsilon_{f}$}}  &0.5 & 1    & 2     & 4  & 8  & 16   & 32    \\
\hline
0.5&0.4345&0.6325&0.6985&0.6989&0.7018&\textbf{0.7168}&0.7221\\
1&0.4468&0.6811&0.7001&0.7028&0.7031&0.7172&0.7243\\
2&0.4714&0.7024&0.7046&0.7054&0.7141&0.7201&0.7302\\
4&0.5473&0.7065&0.7075&0.7098&0.7232&0.7313&0.7355\\
8&0.6397&0.7088&0.7125&0.7228&0.7253&0.7352&\textbf{0.7445}\\
16&0.6859&0.7136&0.7149&0.7255&0.7263&\textbf{0.749}&\textbf{0.7507}\\
32&0.6889&0.7189&0.7289&0.7348&0.735&\textbf{0.7512}&\textbf{0.7543}\\

\bottomrule
\toprule
\multicolumn{8}{|c|}{Deep Tuning \cite{jia2022visual}} \\
\hline
                               \multicolumn{1}{|c|}{\diagbox{$\epsilon_{p}$}{$Acc$}{$\epsilon_{f}$}}  &0.5 & 1    & 2     & 4  & 8  & 16   & 32     \\
\hline
0.5&0.6664&0.6865&0.7012&0.7048&0.7091&0.7109&\textbf{0.7259}\\
1&0.6969&0.7099&0.7142&\textbf{0.7188}&0.7195&\textbf{0.7261}&\textbf{0.7319}\\
2&0.6999&0.7108&0.7201&0.7212&0.7269&\textbf{0.7336}&\textbf{0.736}\\
4&0.7001&\textbf{0.7263}&0.7271&0.7285&\textbf{0.7329}&\textbf{0.7362}&\textbf{0.7385}\\
8&0.7132&0.7275&\textbf{0.7319}&\textbf{0.735}&\textbf{0.7362}&\textbf{0.7429}&0.7432\\
16&0.7182&0.7332&\textbf{0.7359}&\textbf{0.7375}&\textbf{0.7399}&0.7438&0.7445\\
32&0.7239&0.7362&0.7378&0.7409&0.743&0.746&0.75\\

\bottomrule
\toprule
\multicolumn{8}{|c|}{Adapter \cite{houlsby2019parameter}} \\
\hline
                               \multicolumn{1}{|c|}{\diagbox{$\epsilon_{p}$}{$Acc$}{$\epsilon_{f}$}}  &0.5 & 1    & 2     & 4  & 8  & 16   & 32    \\
\hline
0.5&\textbf{0.6985}& \textbf{0.7086}& \textbf{0.7105}& \textbf{0.7112}&\textbf{0.7131}&0.7162&0.7213\\
1&\textbf{0.7044}& \textbf{0.7143}&\textbf{0.7172}&0.7185&\textbf{0.7196}&0.7252&0.7259\\
2&\textbf{0.7158}&\textbf{0.7246}&\textbf{0.7248}&\textbf{0.7261}&\textbf{0.7273}&0.7288&0.7335\\
4&\textbf{0.7175}&0.7261&\textbf{0.7278}&\textbf{0.7291}&0.7298&0.7308&0.7336\\
8&\textbf{0.7209}&\textbf{0.7279}&0.7292&0.7335&0.7345&0.7345&0.7366\\
16&\textbf{0.7288}&\textbf{0.7352}&0.7358&0.7359&0.7366&0.7376&0.7436\\
32&\textbf{0.7368}&\textbf{0.7386}&\textbf{0.7435}&\textbf{0.7452}&\textbf{0.7453}&0.747&0.7475\\

\bottomrule
\toprule
\multicolumn{8}{|c|}{LoRA \cite{hu2021lora}} \\
\hline
                               \multicolumn{1}{|c|}{\diagbox{$\epsilon_{p}$}{$Acc$}{$\epsilon_{f}$}}  &0.5 & 1    & 2     & 4  & 8  & 16   & 32     \\
\hline
0.5&0.6754&0.6772&0.6996&0.7008&0.7096&0.7108&0.7116\\
1&0.6791&0.7048&0.7113&0.7116&0.7155&0.7182&0.7202\\
2&0.6871&0.7155&0.7171&0.7172&0.7181&0.7209&0.7241\\
4&0.6966&0.7178&0.7181&0.7196&0.7198&0.7256&0.7256\\
8&0.7203&0.7215&0.7216&0.7232&0.7258&0.7266&0.7318\\
16&0.7231&0.7253&0.7269&0.7271&0.7292&0.7295&0.7336\\
32&0.7272&0.7293&0.7348&0.737&0.7373&0.7463&0.7472\\

\bottomrule
\toprule
\multicolumn{8}{|c|}{Shallow Tuning \cite{jia2022visual}} \\
\hline
                               \multicolumn{1}{|c|}{\diagbox{$\epsilon_{p}$}{$Acc$}{$\epsilon_{f}$}}  &0.5 & 1    & 2     & 4  & 8  & 16   & 32     \\
\hline
0.5&0.5019&0.5879&0.588&0.6363&0.6801&0.6922&0.6962\\
1&0.6348&0.6959&0.7039&0.7094&0.7095&0.7095&0.7143\\
2&0.6514&0.6984&0.7098&0.7162&0.7173&0.7175&0.7183\\
4&0.7009&0.7046&0.7173&0.7192&0.7203&0.7248&0.7282\\
8&0.7142&0.7188&0.7228&0.7275&0.7306&0.7313&0.7333\\
16&0.7263&0.7269&0.7272&0.7335&0.7342&0.7388&0.7395\\
32&0.736&0.7382&0.7392&0.7409&0.7445&0.7449&0.7452\\

\bottomrule
\toprule
\multicolumn{8}{|c|}{Train from Scratch} \\
\hline
                               $\epsilon$  &0.5 & 1    & 2     & 4  & 8  & 16   & 32     \\
\hline
$Acc$&0.633 & 0.6889 & 0.6998 & 0.7002 & 0.7008 & 0.7011 & 0.7099  \\
                              
\hline

\bottomrule
\toprule
\multicolumn{8}{|c|}{Zero-shot Inference} \\
\hline
                               $\epsilon$  &0.5 & 1    & 2     & 4  & 8  & 16   & 32     \\
\hline
$Acc$      & 0.6471 & 0.6604 & 0.6682& 0.6711 & 0.6814 & 0.7016 & 0.7098
  \\
                              
\hline

\end{tabular}
}
\label{tab2}
\end{table}

Based on the parameter counts in Table \ref{tab1}, we can arrive at the following conclusion. When we make a comparison between the PEFT methods listed in the table and the baseline methods (Full Tuning and Train from Scratch), it becomes evident that all the PEFT approaches have substantially decreased the value of $N$ by at least $\ \frac{206,193-4,408}{206,193}=\textbf{97.86\%}$. To delve into the specifics, it's worth highlighting that LoRA and Adapter emerge as the most parameter-efficient alternatives, exhibiting a remarkable reduction in $N$ by $\frac{206,193 - 1,424}{206,193}=\textbf{99.3\%}$.

\textbf{Testing Accuracy.}  In our endeavor to evaluate the performance of all PEFT methods against the baseline, the TabTransformer model underwent a two-step procedure. Initially, it was subjected to pretraining on the ACSIncome dataset sourced from California (195,665 samples in total), followed by fine-tuning on the ACSIncome dataset from Indiana (35,022 samples in total). The entire process of pretraining and finetuning is performed using DP-SGD. For evaluation, we randomly split $20\%$ of ACSIncome-Indiana as the test set. Furthermore, we opted for the utilization of the accuracy metric, denoted as $Accuracy$ ($Acc$), as the primary evaluation criterion for assessing the TabTransformer's ability to predict whether an individual's annual income exceeds \$50,000. The detailed results are shown in Table \ref{tab2}.

Based on the findings presented in Table \ref{tab2} and Table \ref{tab1}, we can infer the following conclusions. Firstly, all the PEFT methods demonstrate comparable $Acc$ when compared to the baselines. For example, when $\epsilon_p, \epsilon_f$ are both set to $32$, the $Acc$ of Deep Tuning, Adapter, LoRA, and Shallow Tuning is $0.75, 0.7475, 0.7472,\text{ and } 0.7452$, respectively. Meanwhile, when $\epsilon=32$, the $Acc$ of Train from Scratch and Zero-shot Inference are $0.7099$ and $0.7098$, respectively. These values suggest that when compared to Train from Scratch and Zero-shot Inference, the PEFT techniques increase the $Acc$ by at least $\textbf{4.7\%}$. Moreover, the $Acc$ of PEFT is not significantly lagging behind that of Full Tuning. Hence, to sum up, PEFT techniques achieve excellent levels of accuracy ($Acc$) while demonstrating a remarkably high degree of parameter efficiency. Secondly, PEFT methods exhibit a robust tolerance to low values of $\epsilon$ compared to Full Tuning. which indicates that PEFT can ensure a higher level of privacy than Full Tuning. For example, when $(\epsilon_p, \epsilon_f)=(32, 0.5)$, the $Acc$ of Deep Tuning, Adapter, LoRA, and Shallow Tuning are $0.7239, 0.7368, 0.7272, 0.736$, respectively, while the $Acc$ of Full Tuning is $0.6889$. Hence, when $(\epsilon_p, \epsilon_f)=(32,0.5)$, the $Acc$ of PEFT is at least $\textbf{3.5\%}$ higher than Full Tuning. Third,  we find that Deep Tuning and Adapter outperform other methods in most cases. Considering Adapter is more parameter-efficient than Deep Tuning as shown in Table \ref{tab1}, we can conclude that the Adapter achieves the best trade-off among privacy, accuracy, and parameter efficiency.


\section{Conclusion}
 In this paper, we presented a pilot study exploring the benefits of combining 
 differentially private pre-training and parameter-efficient fine-tuning (PEFT) for TabTransfomrers with a variety of fine-tuning methods, including Adapter \cite{houlsby2019parameter}, LoRA \cite{hu2021lora}, Deep/Shallow Tuning \cite{jia2022visual}.
We conducted extensive experiments on the ACSIncome dataset
with different configurations. The results in Table \ref{tab1} indicate that the number of trainable parameters of PEFT techniques reduces at least $\textbf{97.86\%}$ compared to baselines. The results in Table \ref{tab2} show that the accuracy of PEFT methods outperforms baselines in most cases. Hence, compared to three baselines which are either parameter-consuming or ineffective, PEFT techniques achieve a significantly improved trade-off among privacy, accuracy, and parameter efficiency. We also find that Adapter is the most optimal setting for PEFT in this setting. Our study uncovers the unexplored benefits and provides new insights into applying PEPT on differentially private pre-trained TabTransformer for differentially private transfer learning. 

\bibliographystyle{unsrt}
\end{document}